\documentclass[10pt,twocolumn,letterpaper]{article}
\pdfoutput=1

\usepackage{cvpr}
\usepackage{times}
\usepackage{epsfig}
\usepackage{graphicx}
\usepackage{amsmath}
\usepackage{amssymb}
\usepackage{multirow}

\usepackage[pagebackref=true,breaklinks=true,letterpaper=true,colorlinks,bookmarks=false]{hyperref}

\cvprfinalcopy

\def\xhb{\hat{\mathbf{x}}}
\def\yhb{\hat{\mathbf{y}}}
\def\zhb{\hat{\mathbf{z}}}
\def\xb{\mathbf{x}}
\def\yb{\mathbf{y}}
\def\zb{\mathbf{z}}
\def\Ab{\mathbf{A}}
\def\Cb{\mathbf{C}}
\def\Kb{\mathbf{K}}
\newcommand{\fc}[1]{\texttt{fc{#1}}}
\newcommand{\conv}[1]{\texttt{conv{#1}}}

\newcommand{\smallsec}[1]{\textbf{#1:}}

\ifcvprfinal\fi
\begin{document}

\title{Predictive-Corrective Networks for Action Detection}

\author{Achal Dave \hspace{1em} Olga Russakovsky \hspace{1em} Deva Ramanan \\
Carnegie Mellon University \\
}

\maketitle

\begin{abstract}
While deep feature learning has revolutionized techniques for static-image understanding, the same does not quite hold for video processing. Architectures and optimization techniques used for video are largely based off those for static images, potentially underutilizing rich video information. In this work, we rethink both the underlying network architecture and the stochastic learning paradigm for temporal data. To do so, we draw inspiration from classic theory on linear dynamic systems for modeling time series. By extending such models to include nonlinear mappings, we derive a series of novel recurrent neural networks that sequentially make top-down {\bf predictions} about the future and then {\bf correct} those predictions with bottom-up observations. Predictive-corrective networks have a number of desirable properties: (1) they can adaptively focus computation on ``surprising'' frames where predictions require large corrections, (2) they simplify learning in that only ``residual-like'' corrective terms need to be learned over time and (3) they naturally decorrelate an input data stream in a hierarchical fashion, producing a more reliable signal for learning at each layer of a network. We provide an extensive analysis of our lightweight and interpretable framework, and demonstrate that our model is competitive with the two-stream network on three challenging datasets without the need for computationally expensive optical flow.
\end{abstract}
\section{Introduction}

Computer vision is undergoing a period of rapid progress. While the state-of-the-art in image recognition is disruptively increasing, the same does not quite hold for video analysis. Understanding human action in videos, for example, largely remains an unsolved, open problem. Despite a considerable amount of effort, CNN-based features do not yet significantly outperform their hand-designed counterparts for human action understanding~\cite{youtube8M,wang2013action}. We believe that one reason is that many architectures and optimization techniques used for video have largely been inspired by those for static images (so-called ``two-stream'' models~\cite{simonyan2014two,wang2015towards,wang2016temporal,feichtenhofer2016convolutional}), though notable exceptions that directly process spatio-temporal volumes exist~\cite{tran2015learning,varol2016long}. In terms of benchmark results, two-stream models currently outperform the latter, perhaps because of the large computational demands of processing spatio-temporal volumes.

\begin{figure}[t]
\centering
\includegraphics[width=\linewidth]{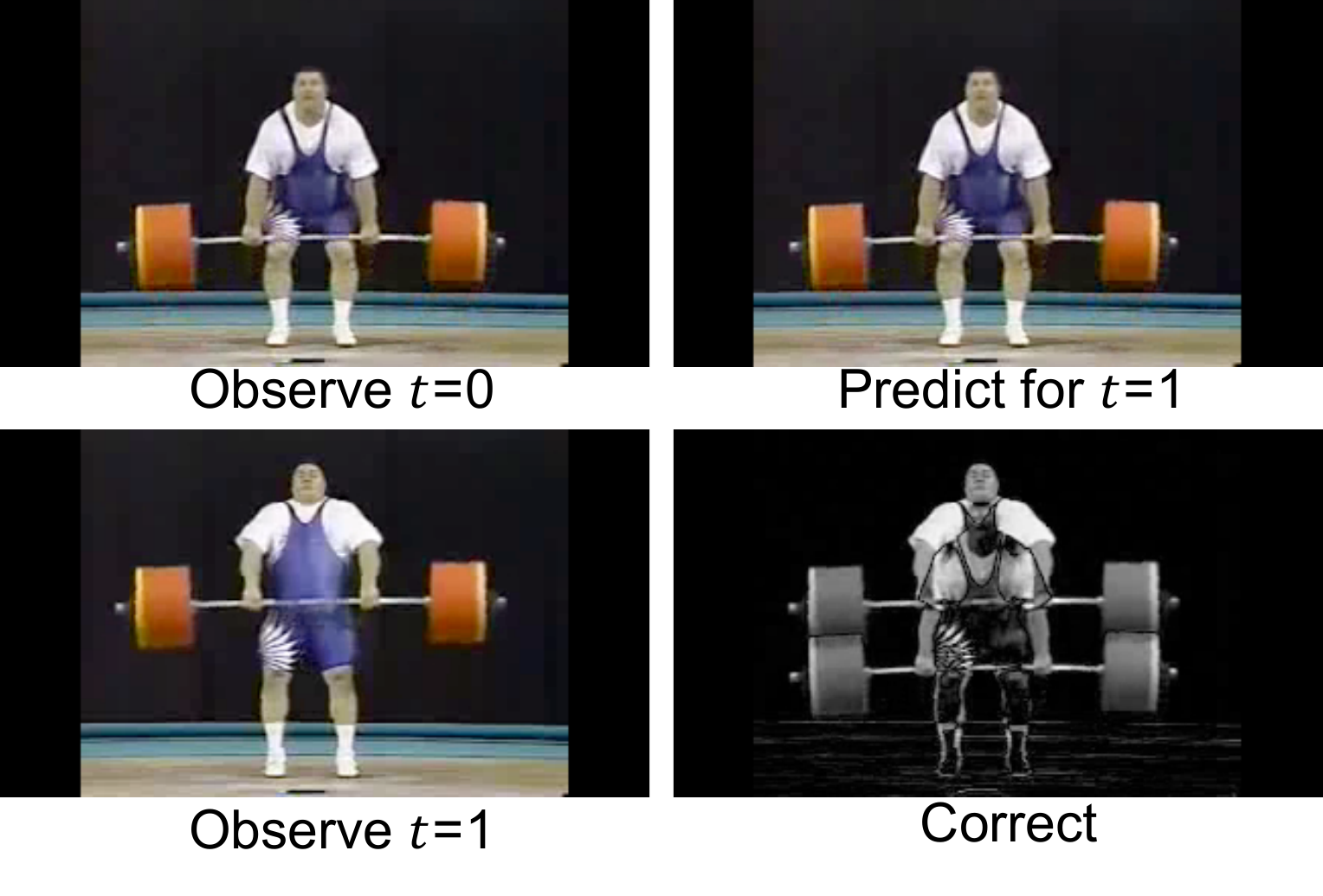}
   \caption{ Our model first \textbf{predicts} the future and then updates its predictions with \textbf{corrections} from observing subsequent frames.}
\label{fig:splash}
\end{figure}

\smallsec{Recurrent models} An attractive solution to the above are {\em state-based} models that implicitly process large spatio-temporal volumes by maintaining a hidden state over time, rather than processing the whole block at once. Classic temporal models based on Hidden Markov Models (HMMs) or Kalman filters do exactly this. Their counterpart in the world of neural nets would be recurrent models. While an active area of research in the context of language~\cite{venugopalan2015sequence,donahue2015long}, they are relatively less explored for video-based feature learning (with the important exceptions of~\cite{yeung2016end,yue2015beyond,srivastava15_unsup_video}). We posit that one reason could be the difficulty of stream-based training with existing SGD solvers. Temporal data streams are highly correlated, while most solvers rely heavily on uncorrelated {\em i.i.d.} data for efficient training~\cite{pearlmutter1995gradient}. Typical methods for ensuring uncorrelated data (such as random data permutations) would remove the very temporal structure that we are trying to exploit!

\smallsec{Our approach} We rethink both the underlying network architecture and stochastic learning paradigm. To do so, we draw inspiration from classic theory on linear dynamic systems for time-series learning models. By extending such iconic models to include nonlinear hierarchical mappings, we derive a series of novel recurrent neural networks that work by making top-down {\em predictions} about the future and {\em correct} those predictions with bottom-up observations (Fig.~\ref{fig:splash}). Just as {\em encoder-decoder} architectures allow for neural networks to incorporate insights from clustering and sparse-coding~\cite{hinton1994autoencoders}, {\em predictive-corrective} architectures allow them to incorporate insights from time-series analysis: (1) adaptively focus computation on ``surprising'' frames where predictions require large corrections, (2)  simplify learning in that only ``residual-like'' corrective terms need to be learned over time and (3)  naturally decorrelate an input stream in a hierarchical fashion, producing a more reliable signal for learning at each layer of a network.

\smallsec{Prediction} From a biological perspective, we leverage the insight that the human vision system relies heavily on continuously predicting the future and then focusing on the unexpected~\cite{enns2008s,mereu2014role}. By utilizing the temporal continuity of video we are able to predict future frames in the spirit of~\cite{vondrick2016generating,vae_eccv2016}. This serves two goals: (1) achieves consistency in predicted actions, reducing the chance that a single noisy frame-level prediction changes the model's interpretation of the video, and (2) results in a computationally efficient system, which is critical for real-world video analysis. If no significant changes are observed between frames then the computation burden can be significantly reduced.

\begin{figure}[t]
\centering
\includegraphics[width=\linewidth]{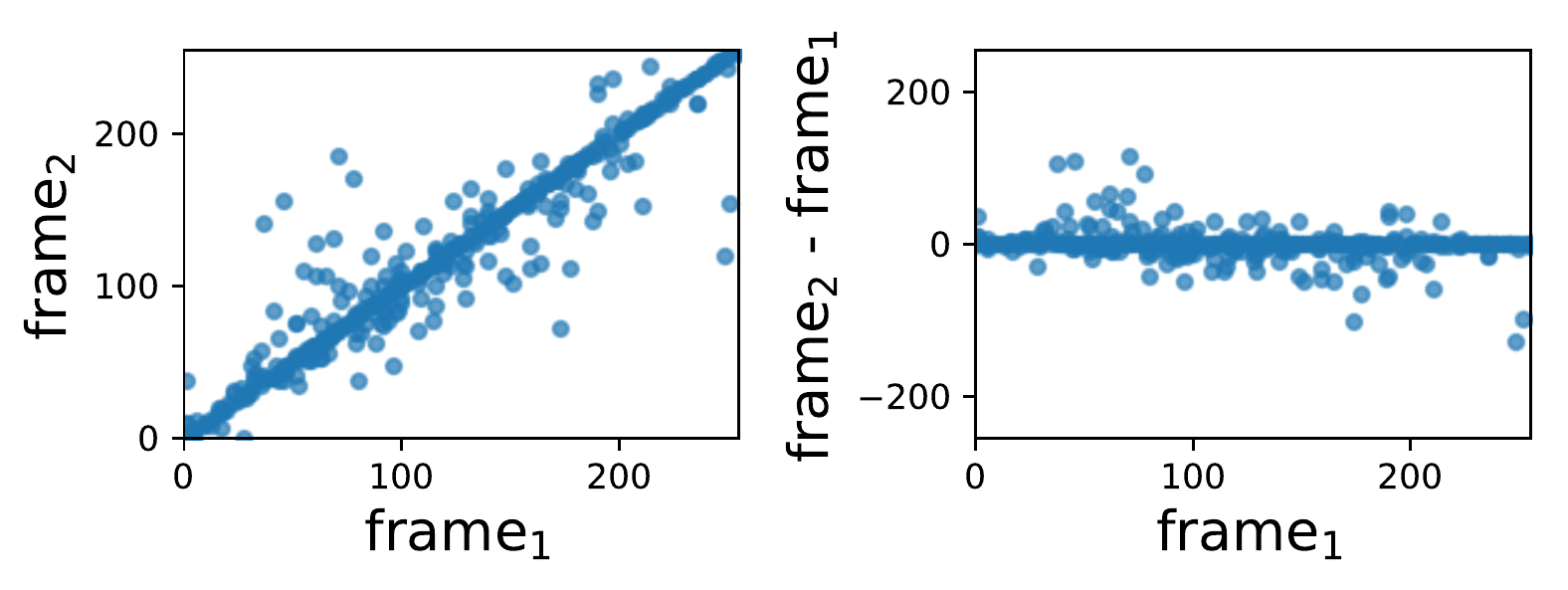}
   \caption{ Every data point corresponds to a single location in two subsequent frames. The x-axis is the pixel intensity at this location in frame$_1$ and the y-axis is the  pixel intensity at this location in frame$_2$ on the left and frame$_2$-frame$_1$ on the right.  \emph{(Left)} Consecutive video frames contain highly correlated information, allowing our model to make accurate and efficient \emph{predictions} about future frames. \emph{(Right)} Explicitly reasoning about frame differences removes correlation and allows the model to focus on reasoning about visual changes by making \emph{corrections} to the predictions.
   }
\label{fig:frame_correlations}
\end{figure}

\smallsec{Correction} Even more importantly,  explicitly modeling appearance predictions allows the model to focus on correcting for novel or unexpected events in the video.  In fine-grained temporal action localization, transitions between actions are commonly signified by only subtle appearance changes. By explicitly focusing on these residual changes our model is able to identify action transitions much more reliably.  Further, from a statistical  perspective, focusing on changes addresses a key challenge in learning from sequential  data: it reduces correlations between consecutive samples, as illustrated in Fig. \ref{fig:frame_correlations}. While consecutive video frames are highly correlated (allowing us to make accurate predictions), \emph{changes} between frames are not, increasing the diversity of samples observed during training.

\smallsec{Contributions} We introduce a lightweight, intuitive and interpretable model for temporal action localization in untrimmed videos. By making predictions about future frames and subsequently correcting its predictions, the model is able to achieve significant improvements in both recognition accuracy and computational efficiency. We demonstrate action localization results on three benchmarks: the standard 20 sports actions of THUMOS~\cite{jiang2014thumos}, the 65 fine-grained actions of MultiTHUMOS~\cite{yeung2015every} and the 157 common everyday actions of Charades~\cite{sigurdsson2016hollywood}. Our model is competitive with the two-stream network~\cite{simonyan2014two} on all three datasets without the need for computationally expensive optical flow. Further, it even (marginally) outperforms the state of the art MultiLSTM model on MultiTHUMOS~\cite{yeung2015every}.

\section{Related Work}

\smallsec{Action recognition} There is a vast literature on action recognition from videos: to name a few, \cite{karpathy2014large,yue2015beyond} explore fusing image-based convolutional networks over time, \cite{simonyan2014two} use RGB pixel information together with optical flow to capture motion, \cite{ji20133d,tran2015learning,varol2016long} extend image-based convolutional networks into 3D convolutional networks that operate on video ``volumes'' consisting of a fixed number of video frames. In contrast to these works, we focus on the more challenging task of temporal action localization.

\smallsec{Temporal action localization} A common way to extend action recognition models to temporal detection is through the sliding windows paradigm~\cite{wang2013action,karamanfast,wangaction,oneata2014lear,yuanadsc}. However, this is both computationally inefficient and prevents the model from leveraging memory over the video. Classical temporal models, on the other hand, can leverage information from the past as well as the future. These models generally rely on chain structured models that admit efficient inference, such as HMMs \cite{shi2011human,hoai2011joint} and CRFs \cite{wang2006hidden}. More recent approaches for reasoning about memory generally focus on Recurrent Neural Networks (RNNs), which sequentially~\cite{yeung2015every} or sporadically~\cite{yeung2016end} process video frames and maintain an explicit memory of the previously observed frames. \cite{koutnik2014clockwork} develop a ``Clockwork RNN'' that maintains memory states that evolve at different speeds while processing a sequence; \cite{shelhamer2016clockwork} extend this model to convolutional networks for semantic segmentation in videos. Our model follows similar intuition for temporal action detection.

\smallsec{Predictive models} It has been shown that leveraging global contextual information can be used to improve image~\cite{Oliva07} or video~\cite{marszalek2009actions} understanding.  Recent work has examined predicting the appearance and semantic content of future video frames~\cite{vondrick2016anticipating,vondrick2016generating,vae_eccv2016,visualdynamics16,finn16unsupervised,lotter2016deep,mathieu2016deep}. Recently Srivastava et al.~\cite{srivastava15_unsup_video} train a recurrent neural network in an encoder-decoder fashion to predict future frames, and demonstrates that the learned video representation improves action recognition accuracy. However, to the best of our knowledge these insights have not been used for designing accurate end-to-end action localization models.

\smallsec{Accelerating learning} Recurrent neural nets are notoriously difficult to train because of the exploding gradients encountered during SGD~\cite{pascanu2013difficulty}.
 We refer the reader to~\cite{bottou2010large} for an excellent introduction to general SGD learning. Though naturally a sequential algorithm that processes one data example at a time, much recent work focuses on mini-batch methods that can exploit parallelism in GPU architectures or clusters~\cite{dean2012large}. One general theme is efficient online approximation of second-order methods~\cite{bordes2009sgd}, which can model correlations between input features. Batch normalization~\cite{ioffe2015batch} computes correlation statistics between samples in a batch, speeding up convergence. Predictive-corrective networks naturally de-correlate batch statistics without needing expensive second-order computations (Fig.~\ref{fig:frame_correlations}).

\smallsec{Interpretable models} Understanding the inner workings of models is important for diagnosing and correcting  mistakes~\cite{parikh2011human,zeiler2014visualizing}. However, despite some recent progress on this front~\cite{karpathy2016visualizing}, recurrent neural networks largely remain a mystery. By introducing a lightweight interpretable recurrent model we aim to gain some insight into the critical components of accurate and efficient video processing.

\section{Predictive-Corrective Model}

Consecutive video frames contain redundant information, which both causes needless extra computation and creates difficulty during training since subsequent samples are highly correlated. In our predictive-corrective model, we remove this redundancy by instead explicitly reasoning about changes between frames. This allows the model to focus on  key visual changes, e.g., corresponding to human motion.

We first provide some intuition motivated by Kalman Filters. Then, we describe a procedural way to apply our model to image-based networks to create recurrent predictive-corrective structures for action detection. The model smoothly updates its memory over consecutive frames through residual corrections based on frame changes, yielding an accurate and efficient framework.

\subsection{Linear Dynamic Systems}
\label{sec:lds}

\smallsec{Setup} Consider a single-shot video sequence that evolves continuously over time. For a video frame at time $t$, let $\xb_t$ denote the underlying semantic representation of the state. For example, on the standard THUMOS dataset~\cite{jiang2014thumos} with 20 sports actions of interest, $\xb_t$ could be a 20-dimensional binary vector indicating the presence or absence of each action within the frame. Instead of a discrete binary vector, we think of $\xb_t$ as a smooth semantic representation: for example, actions can be decomposed into mini muscle motions and $\xb_t$ can correspond to the extent each of these motions is occurring at time $t$. The action detection model is unaware of the underlying state $\xb_t$ but is able to observe the pixel frame appearance $\yb_t$ and is tasked with making an accurate semantic prediction $\xhb_t$ of the state $\xb_t$.

\smallsec{Dynamics} We model the video sequence as a linear dynamic system, which evolves according to
\begin{equation}
\begin{aligned}
\xb_t &= \Ab \xb_{t-1} + \textit{noise} \\
\yb_t &= \Cb \xb_t + \textit{noise}
\end{aligned}
\label{eq:kalman_evolution}
\end{equation}
In other words, the semantic state $\xb_t$ is a noisy linear function of the semantic state at the previous time step $\xb_{t-1}$, and the pixel-level frame appearance $\yb_t$ is a noisy linear function of the underlying semantic action state $\xb_{t}$. This is an imperfect assumption, but intuitively $\xb_t$ can be thought of as action probabilities, $\Ab$ can correspond to the transition matrix between actions, and, if $\xb_t$ is sufficiently high-dimensional, then a linear function can serve as a reasonable approximation of the appearance $\yb_t$.

\smallsec{Kalman filter} Under this linear dynamic system assumption, the posterior estimate of the action state $\xb_{t}$ is:
\begin{equation}
\xhb_{t} = \underbrace{\xhb_{t|t-1}}_{\text{prediction}} + \Kb (\underbrace{\yb_t - \yhb_{t|t-1}}_{\text{correction}}) \label{eq:kalman_update}
\end{equation}
where $\xhb_{t|t-1}$ and $\yhb_{t|t-1}$ are the prior prediction of $\xb_t$ and $\yb_t$ respectively given observations $\yb_1 \dots \yb_{t-1}$ up to previous time steps $t-1$, and $\Kb$ is the Kalman gain matrix. We refer the reader to~\cite{thrun2005probabilistic} for an overview of Kalman filters; for our purposes, we think of $\Kb$ as a learned non-linear function of the difference between actual and predicted frame appearance $\yb_t - \yhb_{t|t-1}$. We analyze Eqn.~\ref{eq:kalman_update} step by step.

\smallsec{State approximation} To make predictions of the semantic action space $\xhb_{t|t-1}$ and of appearance $\yhb_{t|t-1}$, we rely on the fact that the actions and pixel values of a video evolve \textit{slowly} over time \cite{wiskott2002slow}. Using this fact, we can use the previous time step $t-1$ and approximate $\xhb_{t|t-1} \approx \xhb_{t-1}$ with our best prediction of the action state at the previous frame, intuitively saying that the transition matrix between actions in subsequent frames is near-identity. Further, we can assume that the video frame appearance is near constant and approximate $\yhb_{t|t-1} \approx \yb_{t-1}$ with the observed appearance of the previous frame. Eqn.~\ref{eq:kalman_update} now simplifies to:
\begin{equation}
\xhb_{t} = \xhb_{t-1} + g(\yb_t - \yb_{t-1})
\label{eq:pc_block}
\end{equation}
where $g$ is a learned function, which helps compensate for the imperfect assumptions made here.

\smallsec{Learning} What remains is learning the non-linear function $g$ from differences in frame appearance to differences in action state. We call this a \emph{predictive-corrective block} and it forms the basis of our model described below.

\iffalse
Suppose we are interested in predicting some state $\xb_t$ of a system that evolves linearly, guided by the following rule:
\begin{align}
\xb_t = \Ab \xb_{t-1} + \textit{noise}, \label{eq:kalman_evolution}
\end{align}
where $A$ is some linear transformation. At each time step, we observe a noisy measurement of this state, as defined by
\begin{align}
\yb_t = \Cb \xb_t + \textit{noise}
\end{align}
\fi

\subsection{Layered Predictive-Corrective Blocks}
\label{sec:layered_blocks}

\smallsec{Setup} So far we described a general way to predict a hidden state $\xb_t$ given observations $\yb_t$. Instead of thinking of $\yb_t$ as the frame appearance and $\xb_t$ as the semantic action state at time $t$, here we recursively extend our predictive models to capture the hierarchical layers of a deep network: we simply {\em model lower layers as observations that are used to infer the hidden state of higher layers}.  Our model naturally combines hierarchical top-down prediction with hierarchical bottom-up processing prevalent in deep feedforward nets. Let us imagine that layers are computing successively more invariant representations of a video frame, such as activations of parts, objects, actions, etc. We use our model to infer latent parts from image observations, and then treat part activations as observations that can be used to infer objects, and so forth. Let $\zb_t^{l}$ represent the latent activation vector in layer $l$ at time $t$.

\begin{figure}[t]
\centering
\includegraphics[width=\linewidth]{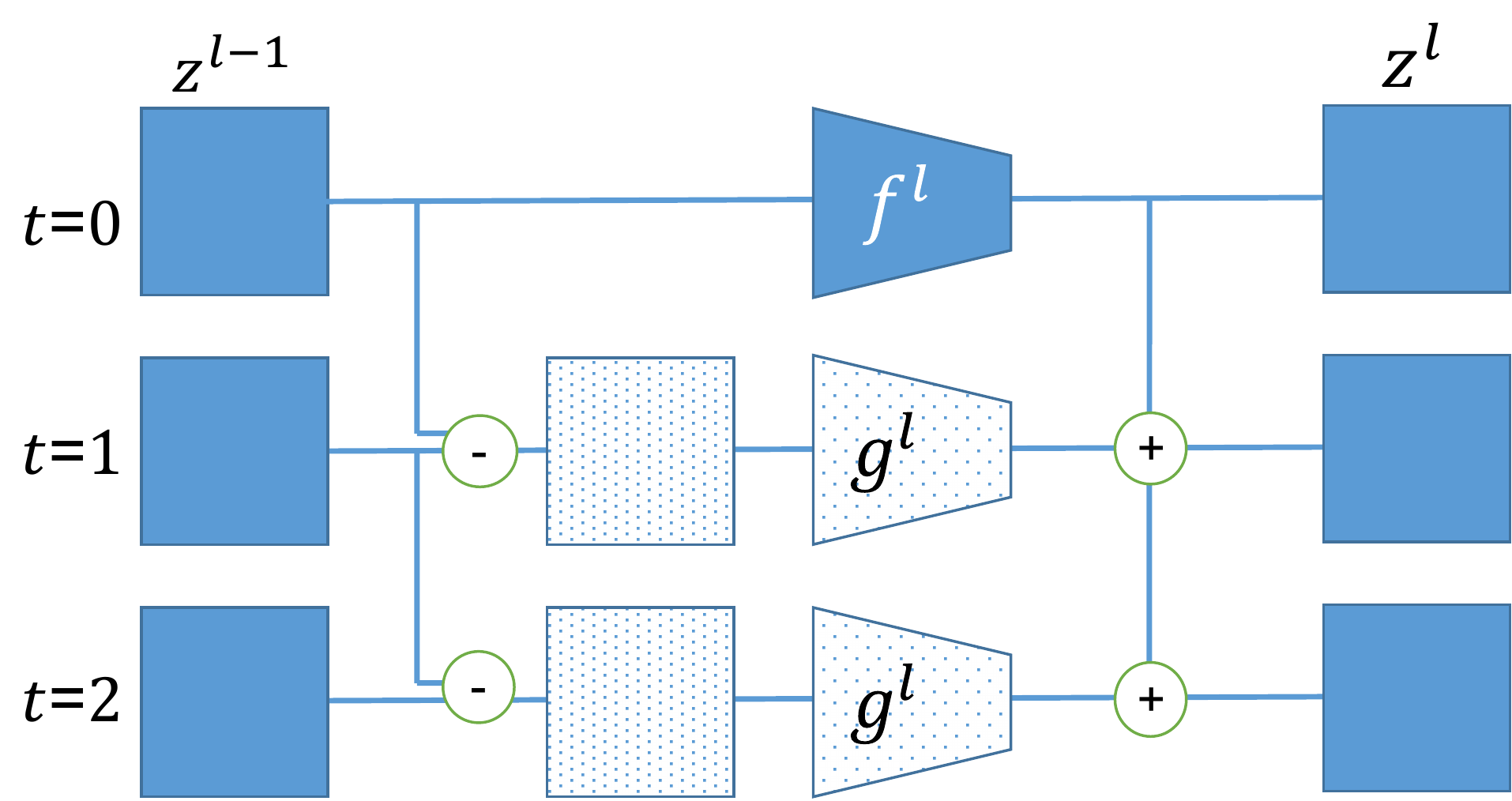}
   \caption{ An instantiation of our predictive-corrective block. The filled and unfilled trapezoids correspond to $f^i$  and $g^l$ respectively. }
\label{fig:pc_block}
\end{figure}

\smallsec{Single layer}
Let us assume $\zb_t^{l}$ evolves over time according to a linear dynamic system that generates observations in the layer below $\zb^{l-1}$. Then the $\zb_t^{l}$ can be predicted as $\zhb_{t}^l$ by observing the temporal evolution of $\zb^{l-1}$ using Eqn.~\ref{eq:pc_block}:
\begin{equation}
\zhb_{t}^l = \zhb_{t-1}^l + g^l(\zb_t^{l-1} - \zb_{t-1}^{l-1})
\label{eq:pc_block_layers}
\end{equation}
There are three things that deserve further discussion. First, the true latent state  $\zb_t^{l}$ can never be observed directly and  we have to rely on predictions  $\zhb_t^{l}$. Second, the base case of the temporal recursion at time $t=0$ needs to be considered. Third, the layer-specific function $g^l$ needs to be learned to predict the evolution of layer $l$ based on the evolution of layer $l-1$.  We now address each of these in order.

\smallsec{Hierarchical model} The latent state of a layer $\zb_t^{l}$ can never be observed except at the lowest layer $l=0$ where $\zb_t^{0}$ is the pixel appearance of the frame. Thus, at each time step $t$ we initialize $\zb_t^{0}$ with the pixel appearance, compute the predicted state $\zhb_{t}^{1}$ using Eqn.~\ref{eq:pc_block_layers}, and use it as the observed $\zb_{t}^{1}$ to compute $\zhb_{t}^{2}$, continuing the layerwise recursion.

\smallsec{Temporal initialization} For the base case of the temporal recursion at time $t=0$ we need a separate convolutional neural network $f$. This network does not consider the evolution of the dynamic system and can be thought of as a simple per-frame (action) recognition model. At the final layer $L$ this network computes the action predictions $\zhb_0^L = f(\zb_0^0)$ from pixel activations $\zb_0^0$; it can also be decomposed layerwise into $\zb_0^l = f^l(\zb_0^{l-1})$. In practice we train $f$ jointly with $g$. Fig.~\ref{fig:pc_block} depicts an instantiation of the system for a single layer $l$, where $f^l$ is used to process the initial frame  and $g^l$ is used to compute sequential updates.

\begin{figure}[t]
\centering
\begin{tabular}{c}
\includegraphics[width=\linewidth]{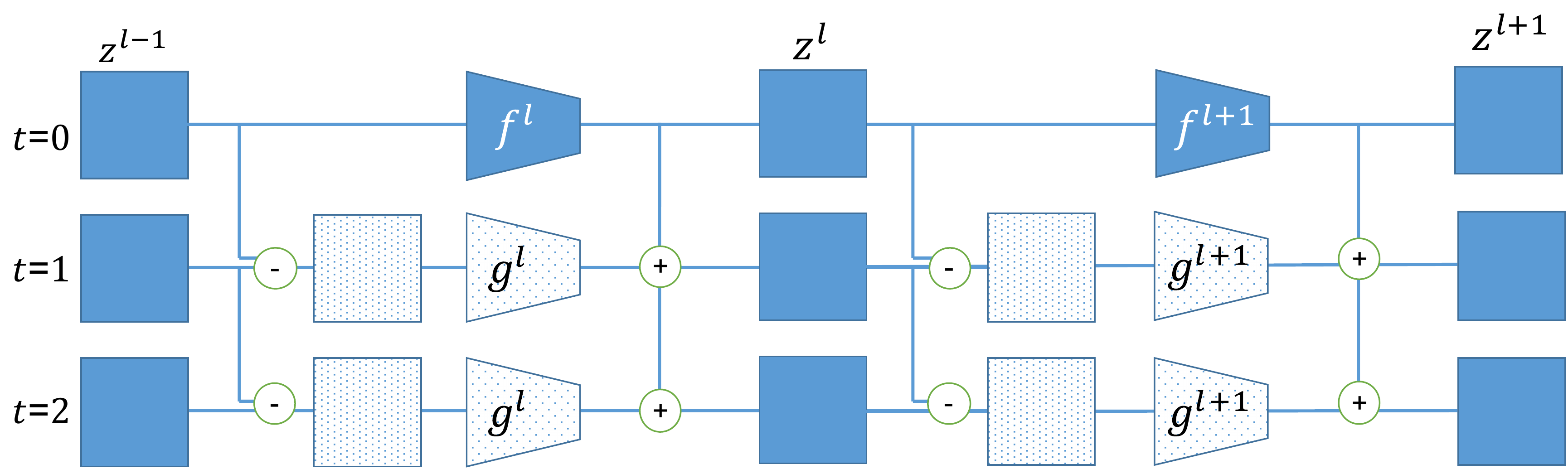} \\
\hline
\includegraphics[width=\linewidth]{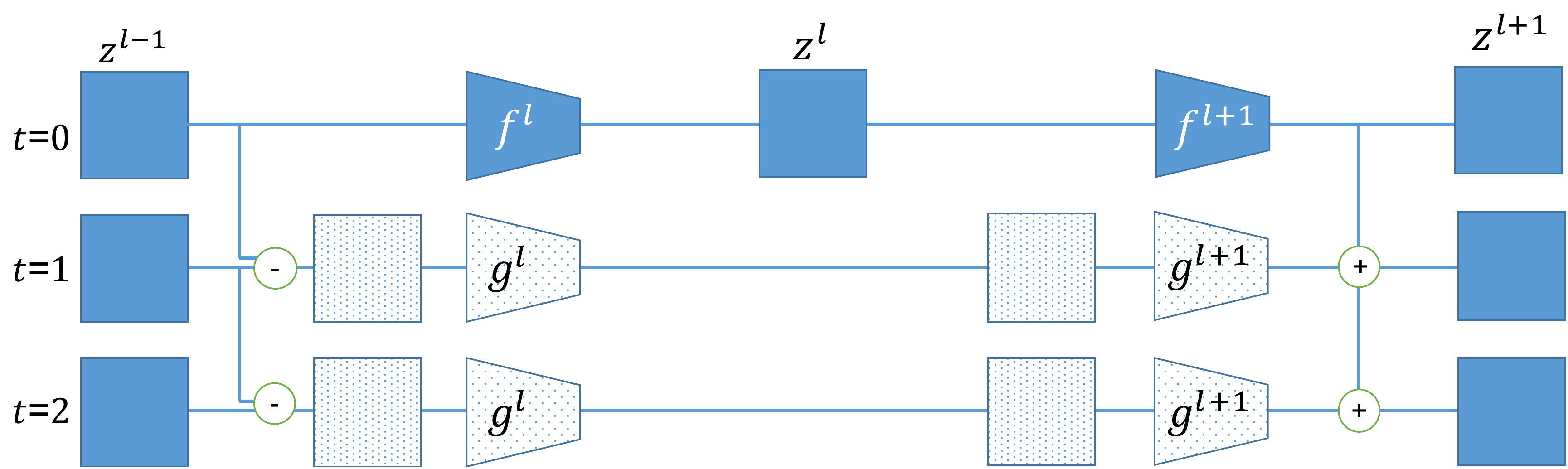}
\end{tabular}
   \caption{ \emph{(Top)} A predictive-corrective block placed at layer $l$ and layer $l+1$. \emph{(Bottom)} An equivalent but simplified network. }
\label{fig:pc_block_stacked}
\end{figure}

\smallsec{Learning} Both the initial-frame function $f$ and the residual function $g$ need to be learned. At any time $t$, we know the pixel frame features $\zb_t^0$ at the zeroth layer of the network and the desired action labels $\zb_t^L$ at the final layer. To compute the action predictions at time $t=0$ we use $\zhb_0^L = f(\zb_0^0)$. To compute the action predictions at time $t\neq 0$ we let $\Delta^l_t \triangleq \zhb^l_t - \zhb^l_{t-1}$ and rewrite the predictive-corrective block equations in Eqn.~\ref{eq:pc_block_layers} as:
\begin{align}
\notag
\Delta^L_t &= g^L(\Delta^{L-1}_t) = g^L(g^{L-1}(\cdots g^1(\Delta^{0}_t))) = g(\zb^0_t - \zb^0_{t-1})
\end{align}
where $\zb^0_t - \zb^0_{t-1}$ is the pixelwise difference between the current and previous frame. Now
we can independently compute $\Delta^L_1, \dots, \Delta^L_t$ using the network $g$
and obtain desired action predictions $\zhb^L_t = \zhb^L_0 + \sum_{i=1}^t\Delta_i^L$ for any time step $t$.  The full system is depicted in Fig.~\ref{fig:pc_block_stacked}. The known action labels $\zb_t^L$ at every time step $t$ provide the training signal for learning the networks $f$ and $g$, and the entire system can be trained end-to-end.

\subsubsection{Connections to prior art}

\indent \indent
\smallsec{Nonlinear Kalman filters} At this point, it is natural to compare our nonlinear dynamic model other nonlinear extensions of Kalman filtering~\cite{haykin2001kalman}. Popular variants include the ``extended'' Kalman filter that repeatedly linearizes a nonlinear model~\cite{ljung1979asymptotic}, and the ``unscented'' Kalman filter that uses particle filtering to model uncertainty under known but nonlinear dynamic function~\cite{julier1997new}. Our work differs in that we assume simple linear dynamics (given by identity mappings), but model the data with complex (nonlinear) hierarchical observation models that are latently-learned from data without hierarchical supervision.

\smallsec{Recurrent networks} We also briefly examine the connection between our framework and a RNN formulation~\cite{yeung2016end,koutnik2014clockwork,yeung2015every}. The update equation for RNN memory is $\xb_t = \sigma(W\yb_t + V\xb_{t-1})$ for input $\yb_t$, non-linearity $\sigma$ and learned weights $W$ and $V$. Similarly, our fully-connected predictive-corrective block in Eqn.~\ref{eq:pc_block} can be written as $\xb_t = \xb_{t-1} + \sigma(W(\yb_t - \yb_{t-1}))$. The key differences are (1) we use the past output $\xb_{t-1}$ in a \emph{linear} fashion, and (2) we maintain the previous input $\yb_{t-1}$  as part of the memory. These imposed constraints are natural for video processing and allow for greater interpretability of the model. Concretely, our memory is simply the convolutional activations at the previous time step, and, thus is as interpretable as the activations of an image-based CNN (using e.g.,~\cite{zeiler2014visualizing}). Second, memory updates are transparent: we clear memory every few frames on re-initialization, and access it only to subtract it from the current convolutional activations, in contrast to the LSTM's more complex update mechanism.

\subsection{Dynamic Computation}
\label{sec:dynamic_computation}

Coming back to our model, so far we discussed the case where layer activations evolve smoothly in a linear dynamic system. However, layer activations between subsequent frames may \emph{not change} at all or may \emph{change too much} to be modeled via smooth updates. Both cases are naturally incorporated into our predictive-corrective model, with the first case additionally yielding computational savings.

\smallsec{Static activations} Layer activations do not change at every time step within a video. This may be because the video depicts a static scene with no moving objects (e.g., in a surveillance camera) or because the frame rate is so high that occasionally subsequent frames appear identical. It may also be the case that while the low-level pixel appearance changes, the higher layers remain static (e.g., a ``face'' neuron that fires regardless of the face's pose or position). Within our model, this leads to $\Delta_t^l = 0$,  eliminating the need for subsequent processing of the corrective block of this frame $t$ for layers $l' > l$ and thus improving efficiency.

\smallsec{Shot changes} On the flip side, occasionally layer activations change so much between subsequent frames that a smooth update is not a reasonable approximation. Then we  ``re-initialize'' by reverting to our  initialization network $f$.

\smallsec{Dynamic updates} Concretely, let $\alpha^l_t$ be an indicator variable representing whether the change in all lower layers $l' < l$ is large enough to warrant corrective computation. Let $\delta^l_t$ be an indicator variable representing whether $\zhb^l_t$ should be reinitialized, either because the change $|\zhb^l_t - \zhb^l_{t-1}|$ is too large or according to a preset layerwise clock rate~\cite{shelhamer2016clockwork,koutnik2014clockwork}. Then, we can rewrite Eqn.~\ref{eq:pc_block_layers} as:
\begin{align}
\zhb^l_t = \begin{cases}
\zhb^l_{t-1} ~&\text{if }  \alpha^{l}_t = 1 \\
f^l(\zhb^{l-1}_t)                                   ~&\text{if } \delta^i_t = 1 \\
\zhb^l_{t-1} + g^l(\zb^{l-1}_t - \zb^{l-1}_{t-1}) ~&\text{else}
\end{cases}
\end{align}
We analyze the effect of dynamic updates on both accuracy and efficiency in our experiments.

\section{Experiments}\label{sec:experiments}
We begin by presenting a detailed analysis of our model with experiments on a validation split of the MultiTHUMOS dataset \cite{yeung2015every} in Sec.~\ref{sec:analysis}. Leveraging this analysis, we then compare the optimal configuration of our predictive-corrective architecture with prior work in Sec.~\ref{sec:test_set_experiments}.

\smallsec{Implementation} For our initial and update models, we use the VGG-16 network architecture \cite{simonyan2014very}. The model is initialized by training on ILSVRC 2016 \cite{russakovsky2015imagenet}. We finetune the model on the per-frame action classification task for all actions, and use these finetuned weights to initialize both the initial and update networks in our model. All of our models are implemented using the Torch \cite{collobert2011torch7} deep learning framework. We will release source code, including hyperparameters and validation splits, for training and evaluating our models. For all of our experiments, we work with frames extracted from the videos at 10 frames per second. Each frame is resized to 256x256 pixels, and we take random crops of 224x224 for each frame.

\subsection{Predictive-Corrective Model Analysis}
\label{sec:analysis}

To analyze the contributions from our proposed approach, we first compare a simple configuration of our approach to baseline models (Sec.~\ref{sec:ablation_baselines}). Next, we evaluate the trade-off of accuracy and efficiency in our framework (Sec.~\ref{sec:reinitialization}). Finally, we consider different model architectures by varying the placement of the predictive-corrective block in the VGG-16 architecture (Sec.~\ref{sec:variations}).

\smallsec{Setup} MultiTHUMOS~\cite{yeung2015every} contains 65 fine-grained action annotations on videos from the THUMOS 2014 dataset~\cite{jiang2014thumos}, which contain 2,765 trimmed training videos, 200 untrimmed training videos, and 213 untrimmed test videos. Of the 200 untrimmed training videos, we select 40 for validation, on which we report experiments below. We evaluate our predictions with per-frame\footnotemark~mean average precision (mAP).~\cite{yeung2015every}

 \footnotetext{Action detection accuracy may also be reported as mAP at specified intersection-over-union (IOU) thresholds, as in \cite{shou2016temporal}. However, this requires post-processing predictions to generate action instances, and we choose not to do that as to not complicate our analysis.}

\subsubsection{Comparison with baselines}
\label{sec:ablation_baselines}
\begin{figure}[t]
\centering
\includegraphics[width=\linewidth]{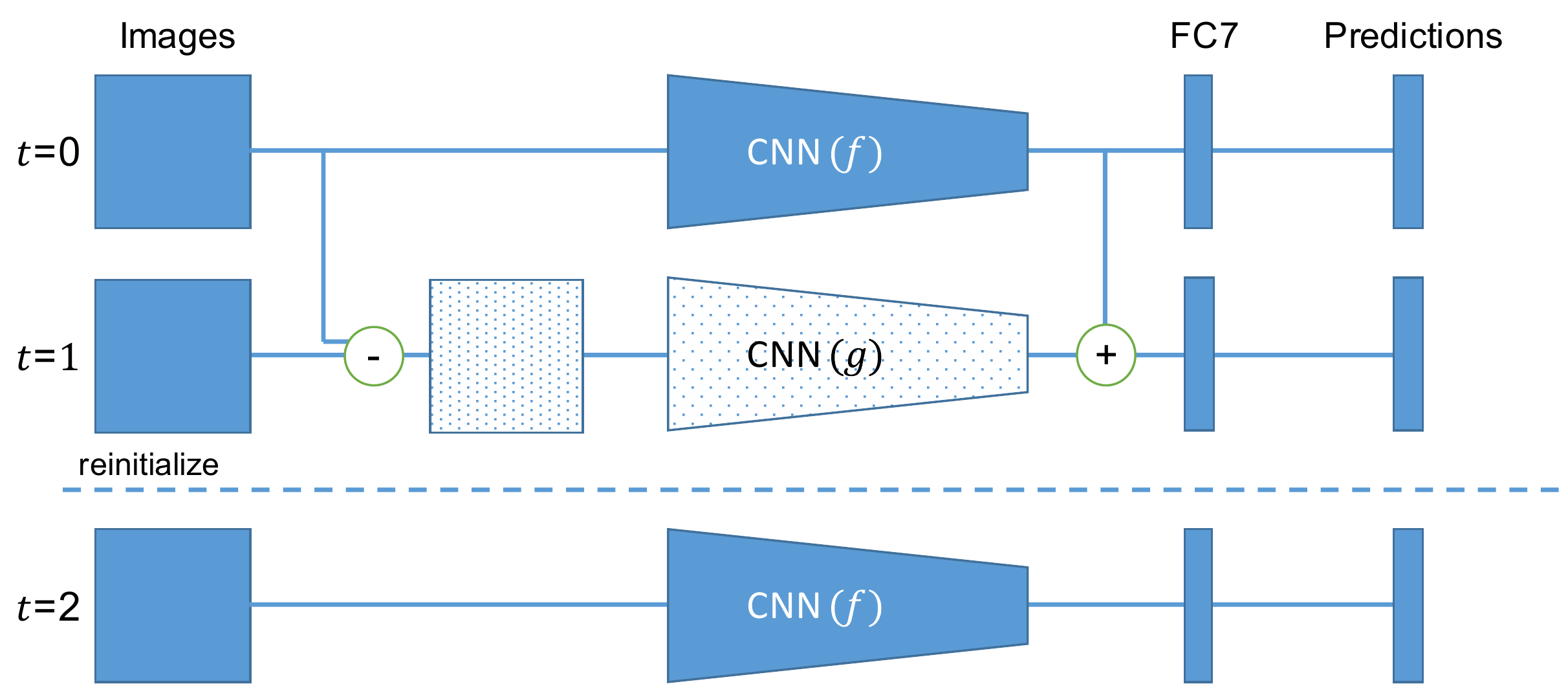}
   \caption{ An instantiation of our predictive-corrective network: a predictive-corrective block  between the input images and the \texttt{fc7} layer with a re-initialization rate of 2 frames. }
\label{fig:fc7_model_reinit_2}
\end{figure}

\indent \indent
\smallsec{Setup} We examine a simple variant of our model: the predictive-corrective block at the \texttt{fc7} layer, which uses frame-level corrections to update \texttt{fc7} activations. In this case, the initial function $f$ and the update function $g$ consist of the layers in VGG-16 up to \texttt{fc7}. Fig.~\ref{fig:fc7_model_reinit_2} shows an instantiation of this with a reinitialization rate of 2. Here we consider the model with a reinitializion rate of 4 frames.

\smallsec{Baselines} We compare our model against the performance of baseline models that do not use our predictive-corrective blocks. To this end, we evaluate two models. First, we evaluate the single frame model finetuned to predict action labels for each frame of a video. Second, we consider a model similar to the \textit{late fusion} model of \cite{karpathy2014large} (or the \textit{late pooling} model of \cite{yue2015beyond}). It takes as input 4 frames (3 from previous time steps plus the current frame) and average pools their \texttt{fc7} activations before making a prediction of the action occurring at the current time step. When training this model we tie the weights corresponding to the three frames, which we found to empirically perform better than leaving all untied or tying all four branches together.

\begin{table}
\centering
\begin{tabular}{|l|c|c|}
\hline
Method                    & MultiTHUMOS mAP       \\ \hline \hline
Single-frame RGB          & 25.1            \\
4-frame late fusion       & 25.3            \\
\hline
Predictive-corrective (our) & {\bf 26.9} \\ \hline
\end{tabular}
\caption{ Our predictive-corrective model outperforms both baselines. (Per-frame mAP on MultiTHUMOS validation set.)}
\label{table:baselines_thumos}
\end{table}

\smallsec{Results} Table~\ref{table:baselines_thumos} reports the results. These baselines explore the contribution of naive temporal information to our performance. While incorporating these cues provide a small $0.2\%$ boosts over the baseline ($25.1\%$ mAP for single-frame vs $25.3\%$ mAP for late fusion), it does not match the performance of our predictive-corrective model. Our model outperforms the single-frame model by $1.8\%$ mAP: from $25.1\%$ mAP of single-frame to $26.9\%$ mAP for ours.
\begin{figure}[t]
\centering
\includegraphics[width=\linewidth]{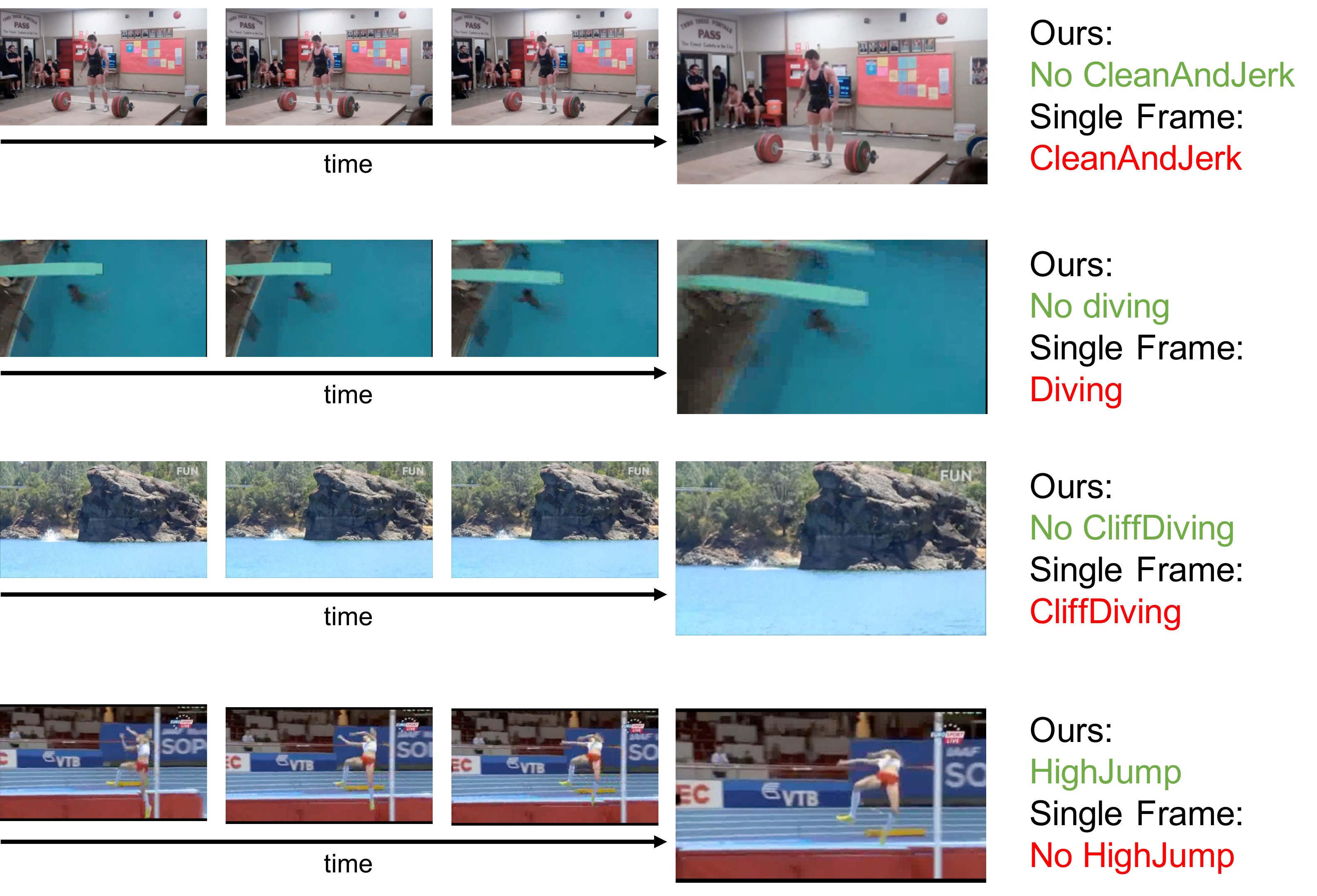}
\caption{ The single-frame model makes predictions based on the overall scene context whereas our predictive-corrective model leverages temporal information from 4 frames to focus on the scene changes and more accurately reason about actions.}
\label{fig:qualitative_samples}
\end{figure}

The single-frame model often relies mostly on the image context when making predictions, yielding many confident false positive predictions as shown in Fig.~\ref{fig:qualitative_samples}. For example, in the top row of Fig.~\ref{fig:qualitative_samples} the single-frame model predicts the ``clean and jerk'' action based on the scene appearance even though the human is not currently performing the action. In contrast, our model is able to effectively use the predictive-corrective block to focus only on the moving parts of the scene and realize that the action is not yet being performed.

\begin{figure}[t]
\centering
\includegraphics[width=0.32\linewidth]{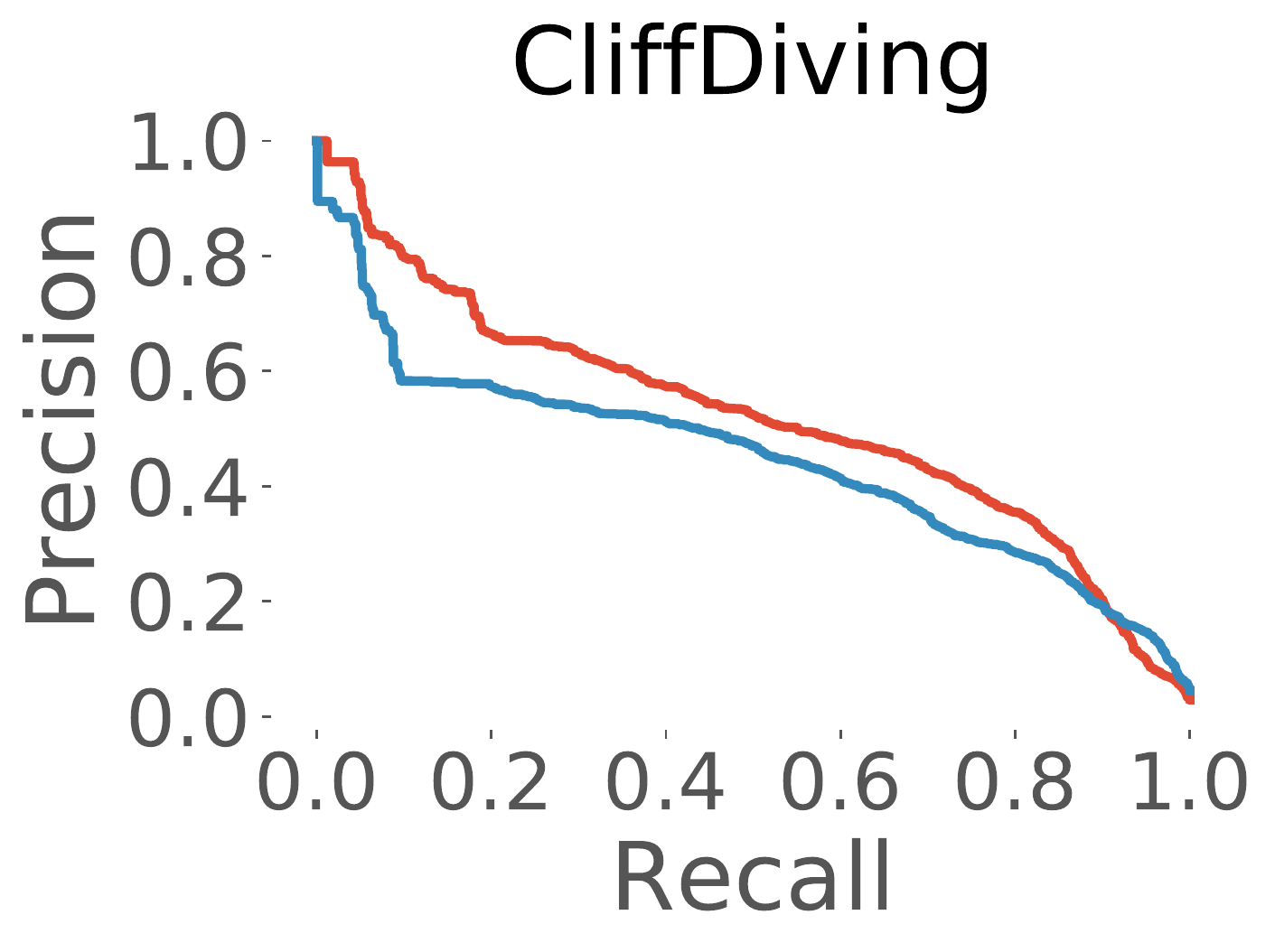}
\includegraphics[width=0.32\linewidth]{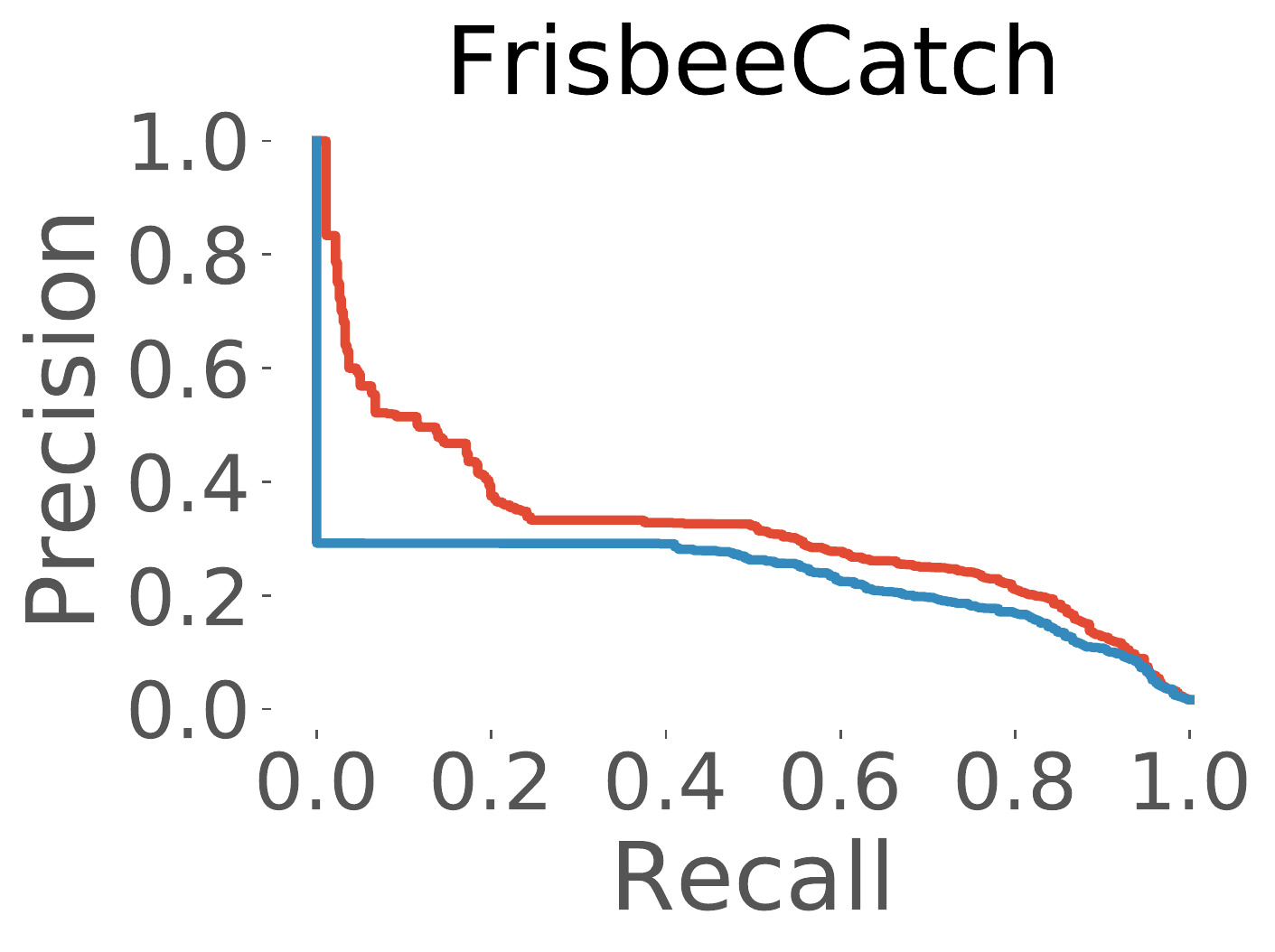}
\includegraphics[width=0.32\linewidth]{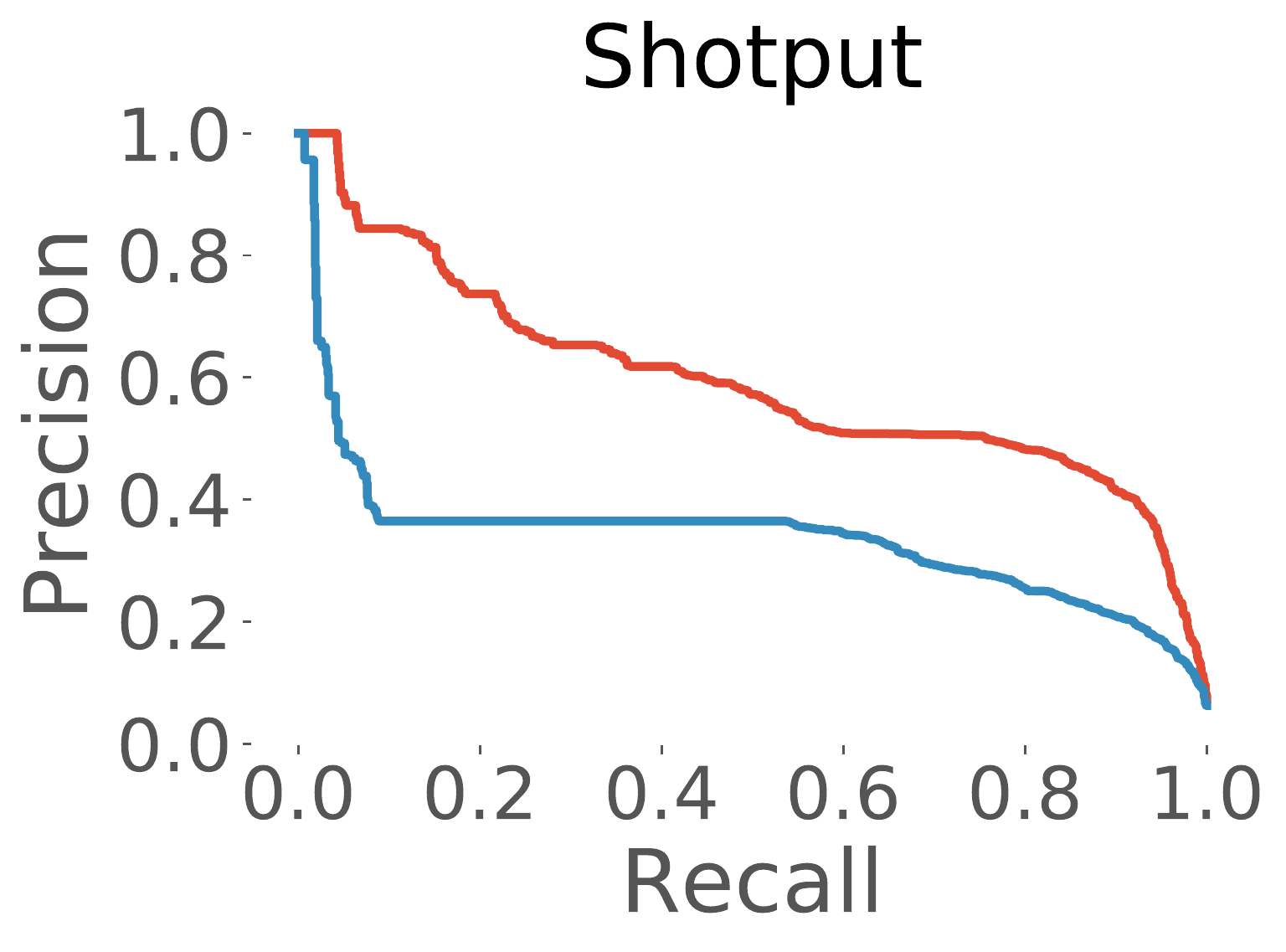}

\caption{ Precision-Recall curves for our model (orange) vs. single-frame (blue). The single-frame model often makes predictions based on context (e.g., ``CliffDiving'' in the presence of a cliff), leading to a lower precision than our model, which leverages temporal information to distinguish actions from context. (Per-frame precision/recall on MultiTHUMOS validation set.)}
\label{fig:pr_plots}
\end{figure}

The precision-recall curves in Fig.~\ref{fig:pr_plots} verify this intuition. The single-frame model consistently suffers from low precision as a result of making many false positive predictions.

\subsubsection{Test-time reinitialization}
\label{sec:reinitialization}

One advantage of our model is that it can be reinitialized dynamically at test time, as described in Sec.~\ref{sec:dynamic_computation}. We have seen results with varying training re-initialization rates in Table~\ref{table:block_placements}. However, these models can be applied in a different setting at test time. This can be useful, for example, if our training data contains videos with many shot changes, but we are interested in evaluating on smooth videos.

\smallsec{Static reinitialization} We experiment with different \textit{train} and \textit{test} reinitialization rates for our \fc7 predictive-corrective model in Table \ref{table:pc_block_static_reinit} for simplicity. The model trained to reinitialize every 4 frames can successfully reason about the video for up to 8 frames without reinitializing with only a modest drop in mAP, while the model trained on 8 frames can generalize to reasoning for up to 16 frames.

\begin{table}
\centering
\begin{tabular}{|l|c|c|}
\hline
Reinit         & Train Reinit 4      & Train Reinit 8 \\ \hline\hline
Test Reinit 2  & 26.9                & 25.9           \\
Test Reinit 4  & 26.9                & 26.9           \\
Test Reinit 8  & 25.4                & \textbf{27.3}  \\
Test Reinit 16 & 20.0                & 25.9           \\
\hline
\end{tabular}

\caption{Our model is able to reason about the video at test time for longer than it was trained for, with only modest losses in accuracy.  (Per-frame mAP on MultiTHUMOS validation set.)}
\label{table:pc_block_static_reinit}
\end{table}

\begin{table}[t]
\centering
\begin{tabular}{|l|c|c|}
\hline
Configuration           &  mAP  \\ \hline\hline
\conv{53} every 4 & 26.5 \\
\fc7 every 4    & 26.9 \\
\fc8 every 4    & 26.6 \\ \hline
\conv{33} every 1, \fc7 every 4    & \textbf{27.2} \\
\conv{43} every 2, \conv{53} every 4 & 26.6  \\
\conv{53} every 2, \fc7 every 4    & 24.8 \\
\hline
\end{tabular}
\caption{Accuracy of different predictive-corrective architectures. (Per-frame mAP on MultiTHUMOS validation set.)}
\label{table:block_placements}
\end{table}

\smallsec{Dynamic reinitialization} In addition to static reinitialization rates, our model is able to \textit{dynamically} decide when to re-initialize at test time. This allows it to use the corrective model when the video is evolving smoothly, and re-initialize only during big time changes. We implement this by thresholding the corrective term computed in the given frame; if its magnitude is greater than our threshold, we re-initialize the model. To avoid propagating mistakes for long sequences, we require the model to re-initialize at least every 4th frame. We find that by validating over a simple dynamic threshold on the norm of the correction, we can already achieve a small improvement in accuracy from the static reinitialization rate ($27.2\%$ mAP dynamic  vs $26.9\%$ mAP static). This suggests that using more advanced methods such as reinforcement learning to learn dynamic updates may yield further benefits in our frameworks.

\smallsec{Efficiency} Processing videos is computationally challenging due to the heavy redundancies between frames. Our model naturally allows us to avoid unnecessary computation on frames that are mostly redundant. We implement this by discarding frames when the corrective term is below a threshold. We find that we can dynamically discard nearly 50\% of the frames, thus reducing the computational burden by a factor of two, while only slightly dropping performance ($26.7\%$ mAP with processing only half the frames vs $26.9\%$ mAP with processing all frames). Note that this is not the same as randomly discarding frames, as our model still outputs predictions for {\em all} frames.

\subsubsection{Architectural Variations}
\label{sec:variations}

We have so far considered a model with a predictive-corrective block at the \fc7 layer that re-initializes every 4 frames. However, different layers of the network capture different information about the video, and evolve at different rates. We investigate these options to gain deeper insight into the model and into the structure of temporal data.

\smallsec{Single block} We begin by experimenting on models with a single predictive-corrective block. We consider placing the block at different layers in the model other than  \fc7, thus asking the model to focus on more low-level (\conv{53}) or high-level (\texttt{fc8}) visual changes. Table~\ref{table:block_placements} reports the results. We find that placing a predictive corrective block at \fc7 is optimal within the single-block setting. Placing the block at either \conv{53} or \fc8 yields a $0.4\%$ and $0.3\%$ respective reduction in mAP.  Reasoning about higher-level but non-semantic features proves to be the most effective.

\smallsec{Hierarchical blocks} By placing a single predictive-corrective block, we force the entire model to reinitialize its memory at the same rate. We hypothesize that re-initializing at a faster rate may be important, particularly for predictive-corrective blocks placed at lower levels in the network since the low-level visual features change faster than the more semantic \texttt{fc7}. Encouraged by this intuition, we experiment with placing predictive-corrective blocks at multiple layers with different reinitialization rates. We explore a few hierarchical configurations in Table~\ref{table:block_placements}. In particular, the ``\conv{33} every 1, \fc7 every 4'' model can be interpreted as predicting and correcting \conv{33} activations instead of \textit{pixel} values (as the ``\fc7 every 4'' model does), which are less sensitive to noise, brightness changes, and camera motion than raw pixels. Indeed, this model outperforms all other configurations, achieving $27.2\%$ mAP.

\begin{figure}[t]
\centering
\includegraphics[width=\linewidth]{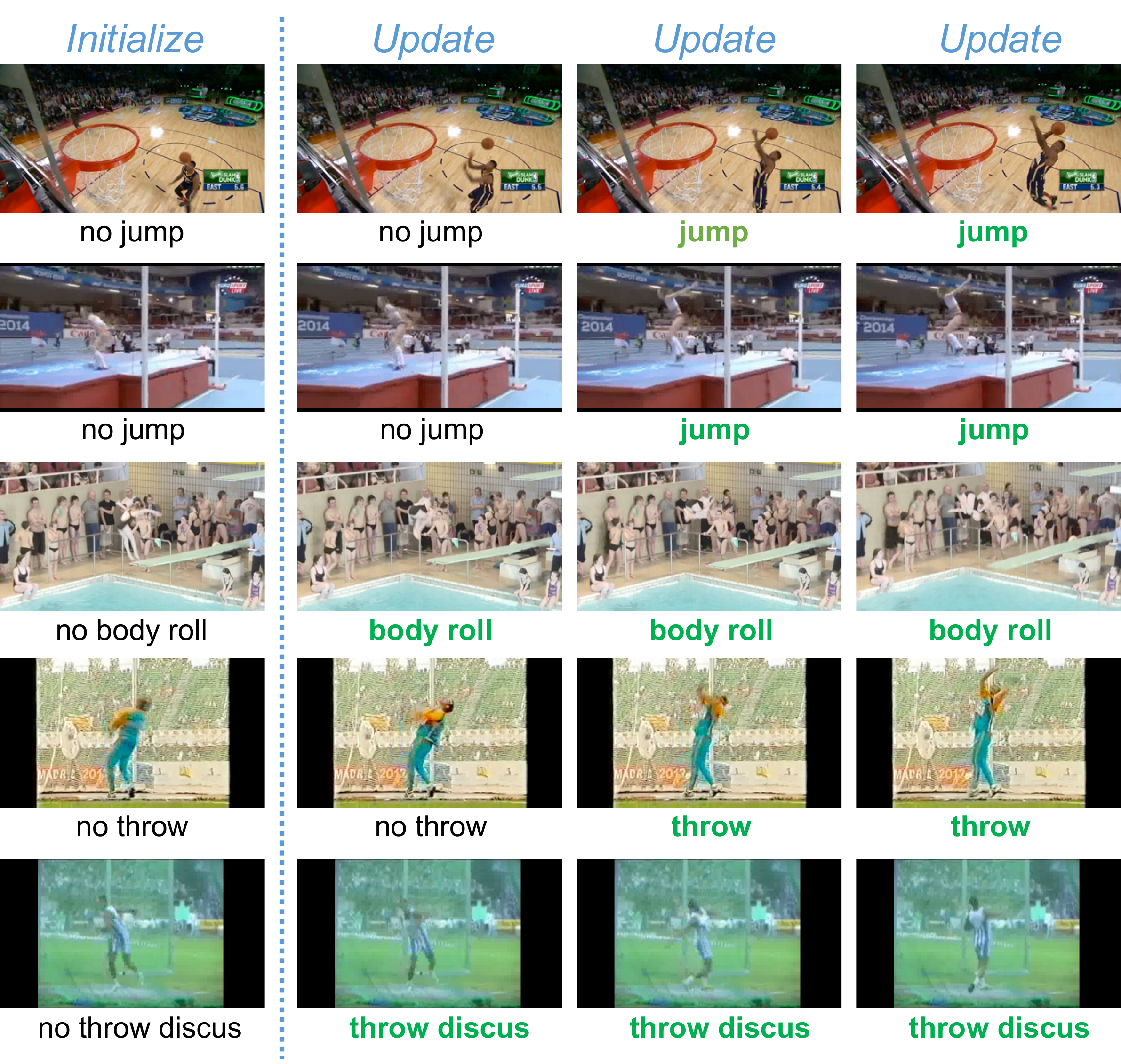}
\caption{ Qualitative results on the MultiTHUMOS validation set. Labels are our model's predictions for each frame. Our model initializes on the first frame and updates using the next three frames. Our update mechanism correctly recognizes the start of actions after initialization, and even corrects errors from initialization (last).}
\label{fig:qualitative_results}
\end{figure}

\smallsec{Effective corrections} We conclude with a qualitative look into the predictions made by our model. In particular, one worry is that the model may be predicting the same action labels across all 4 frames between reinitializations. Fig.~\ref{fig:qualitative_results} shows that this is not the case. The predictive-corrective block is able to successfully notice the changes that occur between frames and update the action predictions. For example, in the first row of
Fig.~\ref{fig:qualitative_results} the ``jump'' action happens 2 frames after reinitialization, and the model successfully corrects its initial prediction.

\subsection{Comparison to Prior Approaches}
\label{sec:test_set_experiments}

Building off our analysis in Sec.~\ref{sec:analysis}, we now evaluate our predictive-corrective model on three challenging benchmarks: MultiTHUMOS~\cite{yeung2015every}, THUMOS~\cite{jiang2014thumos}, and Charades~\cite{sigurdsson2016hollywood}. Table~\ref{table:block_placements} motivates using the hierarchical ``\conv{33} every 1, \fc7 every 4'' architecture; Table~\ref{table:pc_block_static_reinit} demonstrates that  training to reinitialize every 8 frames yields further improvement. Thus we use the ``\conv{33} every 1, \fc7 every 8'' as our predictive-corrective model.

\subsubsection{THUMOS and MultiTHUMOS}
\indent \indent
\smallsec{Setup}  THUMOS~\cite{jiang2014thumos} contains 20 annotated action classes; MultiTHUMOS~\cite{yeung2015every} includes 45 additional action classes annotated on the THUMOS videos. We train models on the training and validation videos for all the MultiTHUMOS actions jointly. We then evaluate on the THUMOS test videos by computing the per-frame mAP over the 20 THUMOS and 65 MultiTHUMOS action classes.

\smallsec{Results} We report results in Table~\ref{table:test_multithumos_thumos_partial_train}. The single-frame model has been shown to be a strong baseline for action recognition, outperforming, e.g., C3D~\cite{tran2015learning} in \cite{sigurdsson2016hollywood}. On MultiTHUMOS our predictive-corrective model not only outperforms the single-frame baseline by $4.3\%$ mAP ($29.7\%$ mAP ours vs $25.4\%$ mAP single-frame), but also compares favorably to the state-of-the art MultiLSTM model~\cite{yeung2015every}.

On THUMOS, our model still outperforms the single-frame model by $4.2\%$ ($38.9\%$ mAP ours vs $34.7\%$ mAP single-frame), but is not yet on par with MultiLSTM. This may be due to the significantly longer actions in THUMOS, which the LSTM-based model can handle better due to a longer (though less interpretable) memory of the video.

At the cost of efficiency, we can further improve our model by running it in a dense sliding window fashion where the model has a 7 frame history when making a prediction for each frame. With this approach, our model achieves $30.8\%$ mAP on MultiTHUMOS (significantly outperforming MultiLSTM's $29.7\%$ mAP) and $40.9\%$ on THUMOS (only $0.4\%$ behind MultiLSTM at $41.3\%$ mAP).

\subsubsection{Charades}
\indent \indent
\smallsec{Setup} Whereas the THUMOS and MultiTHUMOS datasets contain primarily videos of sports actions, the Charades dataset~\cite{sigurdsson2016hollywood} contains videos of common everyday actions performed by people in their homes. The dataset contains 7,986 untrimmed training videos and 1,864 untrimmed test videos, with a total of 157 action classes. This is a significantly more challenging testbed: first, it contains many more actions than MultiTHUMOS, and second, it is constructed so as to decorrelate actions from scenes.

\smallsec{Results} Our model generalizes to this new domain despite the challenges. We report action localization results (following~\cite{sigurdsson2016asynchronous}) in Table~\ref{table:test_charades}. Our predictive-corrective model improves from $7.9\%$ mAP of the single-frame baseline and the $7.7\%$ mAP of the LSTM baseline to $8.9\%$ mAP. Further, our model is able to match the accuracy of the two-stream network, without the need for explicitly computing expensive optical flow.\footnote{For completeness, we note that the model does not yet match state-of-the-art results on the Charades benchmark: e.g., \cite{sigurdsson2016asynchronous} achieves  $12.5\%$ mAP using global cues and post-processing.}

\begin{table}[t]
\centering
\begin{tabular}{|l|c|c|}
\hline
Method                            & MultiTHUMOS   & THUMOS \\ \hline \hline
Single-frame \cite{yeung2015every} & 25.4          & 34.7 \\
Two-Stream\footnotemark \cite{simonyan2014two} & 27.6          & 36.2 \\
Multi-LSTM \cite{yeung2015every}   & \textbf{29.7}          & \textbf{41.3} \\
Predictive-corrective              & \textbf{29.7} & 38.9 \\
\hline
\end{tabular}
\caption{Comparison of our model with prior art. (Per-frame mAP on MultiTHUMOS and THUMOS test sets.)
}
\label{table:test_multithumos_thumos_partial_train}
\end{table}

\footnotetext{The two-stream number is reported from \cite{yeung2015every}, which uses a single optical flow frame for the flow stream.}

\begin{table}[t]
\centering
\begin{tabular}{|l|c|c|}
\hline
Method                   & Charades \\ \hline \hline
Single-frame             & 7.9 \\
LSTM (on RGB)            & 7.7 \\
Two-Stream \cite{charadeswebsite} & \textbf{8.9} \\
Predictive-corrective    & \textbf{8.9} \\
\hline
\end{tabular}
\caption{Comparison of our model with prior work on Charades. Our model matches the accuracy of the two-stream model without using optical flow. (Localization mAP on the Charades test set.)}
\label{table:test_charades}
\end{table}

\section{Conclusions}
We introduced a recurrent predictive-corrective network that maintains an \textit{interpretable} memory that can be dynamically re-initialized. Motivated by Kalman Filters, we exploit redundancies and motion cues within videos to smoothly update our per-frame predictions and intermediate activations within a convolutional network. We perform extensive ablation studies of this model, carefully choosing where to place predictive-corrective blocks, improving accuracy over baselines on the MultiTHUMOS and THUMOS datasets.

\subsubsection*{Acknowledgements}
We would like to thank Rohit Girdhar and James Supan\v ci\v c for reviewing early versions of this paper, and Serena Yeung and Gunnar Sigurdsson for help working with the MultiTHUMOS and Charades datasets. Funding for this research was provided by NSF Grant 1618903, NSF Grant 1208598, and the Intel Science and Technology Center for Visual Cloud Systems (ISTC-VCS).

{
\small
\bibliographystyle{ieee}
\bibliography{main}

\begin{thebibliography}{10}\itemsep=-1pt

\bibitem{youtube8M}
S.~Abu{-}El{-}Haija, N.~Kothari, J.~Lee, P.~Natsev, G.~Toderici,
  B.~Varadarajan, and S.~Vijayanarasimhan.
\newblock Youtube-8m: {A} large-scale video classification benchmark.
\newblock {\em CoRR}, abs/1609.08675, 2016.

\bibitem{bordes2009sgd}
A.~Bordes, L.~Bottou, and P.~Gallinari.
\newblock Sgd-qn: Careful quasi-newton stochastic gradient descent.
\newblock {\em The Journal of Machine Learning Research}, 10:1737--1754, 2009.

\bibitem{bottou2010large}
L.~Bottou.
\newblock Large-scale machine learning with stochastic gradient descent.
\newblock In {\em Proceedings of COMPSTAT'2010}, pages 177--186. Springer,
  2010.

\bibitem{collobert2011torch7}
R.~Collobert, K.~Kavukcuoglu, and C.~Farabet.
\newblock Torch7: A matlab-like environment for machine learning.
\newblock In {\em BigLearn, NIPS Workshop}, number EPFL-CONF-192376, 2011.

\bibitem{dean2012large}
J.~Dean, G.~Corrado, R.~Monga, K.~Chen, M.~Devin, M.~Mao, A.~Senior, P.~Tucker,
  K.~Yang, Q.~V. Le, et~al.
\newblock Large scale distributed deep networks.
\newblock In {\em Advances in Neural Information Processing Systems}, pages
  1223--1231, 2012.

\bibitem{donahue2015long}
J.~Donahue, L.~Anne~Hendricks, S.~Guadarrama, M.~Rohrbach, S.~Venugopalan,
  K.~Saenko, and T.~Darrell.
\newblock Long-term recurrent convolutional networks for visual recognition and
  description.
\newblock In {\em Computer Vision and Pattern Recognition}, 2015.

\bibitem{enns2008s}
J.~T. Enns and A.~Lleras.
\newblock What's next? new evidence for prediction in human vision.
\newblock {\em Trends in cognitive sciences}, 12(9):327--333, 2008.

\bibitem{feichtenhofer2016convolutional}
C.~Feichtenhofer, A.~Pinz, and A.~Zisserman.
\newblock Convolutional two-stream network fusion for video action recognition.
\newblock In {\em Computer Vision and Pattern Recognition}, 2016.

\bibitem{finn16unsupervised}
C.~Finn, I.~Goodfellow, and S.~Levine.
\newblock Unsupervised learning for physical interaction through video
  prediction.
\newblock In {\em NIPS}, 2016.

\bibitem{haykin2001kalman}
S.~S. Haykin et~al.
\newblock {\em Kalman filtering and neural networks}.
\newblock Wiley Online Library, 2001.

\bibitem{hinton1994autoencoders}
G.~E. Hinton and R.~S. Zemel.
\newblock Autoencoders, minimum description length, and helmholtz free energy.
\newblock {\em Neural Information Processing Systems}, 1994.

\bibitem{hoai2011joint}
M.~Hoai, Z.-Z. Lan, and F.~De~la Torre.
\newblock Joint segmentation and classification of human actions in video.
\newblock In {\em Computer Vision and Pattern Recognition (CVPR)}, 2011.

\bibitem{ioffe2015batch}
S.~Ioffe and C.~Szegedy.
\newblock Batch normalization: Accelerating deep network training by reducing
  internal covariate shift.
\newblock In D.~Blei and F.~Bach, editors, {\em Proceedings of the 32nd
  International Conference on Machine Learning (ICML-15)}, pages 448--456. JMLR
  Workshop and Conference Proceedings, 2015.

\bibitem{ji20133d}
S.~Ji, W.~Xu, M.~Yang, and K.~Yu.
\newblock 3d convolutional neural networks for human action recognition.
\newblock {\em IEEE transactions on pattern analysis and machine intelligence},
  35(1):221--231, 2013.

\bibitem{jiang2014thumos}
Y.~Jiang, J.~Liu, A.~R. Zamir, G.~Toderici, I.~Laptev, M.~Shah, and
  R.~Sukthankar.
\newblock Thumos challenge: Action recognition with a large number of classes.
\newblock In {\em ECCV Workshop}, 2014.

\bibitem{julier1997new}
S.~J. Julier and J.~K. Uhlmann.
\newblock New extension of the kalman filter to nonlinear systems.
\newblock In {\em AeroSense'97}, pages 182--193. International Society for
  Optics and Photonics, 1997.

\bibitem{karamanfast}
S.~Karaman, L.~Seidenari, and A.~Del~Bimbo.
\newblock Fast saliency based pooling of fisher encoded dense trajectories.
\newblock In {\em THUMOS'14 challenge entry}, 2014.

\bibitem{karpathy2016visualizing}
A.~Karpathy, J.~Johnson, and L.~Fei-Fei.
\newblock Visualizing and understanding recurrent networks.
\newblock In {\em ICLR Workshop}, 2016.

\bibitem{karpathy2014large}
A.~Karpathy, G.~Toderici, S.~Shetty, T.~Leung, R.~Sukthankar, and L.~Fei-Fei.
\newblock Large-scale video classification with convolutional neural networks.
\newblock In {\em Computer Vision and Pattern Recognition}, 2014.

\bibitem{koutnik2014clockwork}
J.~Koutnik, K.~Greff, F.~Gomez, and J.~Schmidhuber.
\newblock A clockwork rnn.
\newblock In {\em International Conference on Machine Learning}, 2014.

\bibitem{ljung1979asymptotic}
L.~Ljung.
\newblock Asymptotic behavior of the extended kalman filter as a parameter
  estimator for linear systems.
\newblock {\em IEEE Transactions on Automatic Control}, 24(1):36--50, 1979.

\bibitem{lotter2016deep}
W.~Lotter, G.~Kreiman, and D.~Cox.
\newblock Deep predictive coding networks for video prediction and unsupervised
  learning.
\newblock {\em CoRR}, abs/1605.08104, 2016.

\bibitem{marszalek2009actions}
M.~Marszalek, I.~Laptev, and C.~Schmid.
\newblock Actions in context.
\newblock In {\em Computer Vision and Pattern Recognition (CVPR)}, 2009.

\bibitem{mathieu2016deep}
M.~Mathieu, C.~Couprie, and Y.~LeCun.
\newblock Deep multi-scale video prediction beyond mean square error.
\newblock In {\em ICLR}, 2016.

\bibitem{mereu2014role}
S.~Mereu, J.~M. Zacks, C.~A. Kurby, and A.~Lleras.
\newblock The role of prediction in perception: Evidence from interrupted
  visual search.
\newblock {\em Journal of experimental psychology: human perception and
  performance}, 40(4):1372, 2014.

\bibitem{Oliva07}
A.~Oliva and A.~Torralba.
\newblock The role of context in object recognition.
\newblock {\em Trends in Cognitive Sciences}, 11(12), 2007.

\bibitem{oneata2014lear}
D.~Oneata, J.~Verbeek, and C.~Schmid.
\newblock The lear submission at thumos 2014.
\newblock In {\em THUMOS'14 challenge}, 2014.

\bibitem{parikh2011human}
D.~Parikh and C.~Zitnick.
\newblock Human-debugging of machines.
\newblock {\em NIPS WCSSWC}, 2:7, 2011.

\bibitem{pascanu2013difficulty}
R.~Pascanu, T.~Mikolov, and Y.~Bengio.
\newblock On the difficulty of training recurrent neural networks.
\newblock {\em ICML (3)}, 28:1310--1318, 2013.

\bibitem{pearlmutter1995gradient}
B.~A. Pearlmutter.
\newblock Gradient calculations for dynamic recurrent neural networks: A
  survey.
\newblock {\em IEEE Transactions on Neural networks}, 6(5), 1995.

\bibitem{russakovsky2015imagenet}
O.~Russakovsky, J.~Deng, H.~Su, J.~Krause, S.~Satheesh, S.~Ma, Z.~Huang,
  A.~Karpathy, A.~Khosla, M.~Bernstein, et~al.
\newblock Imagenet large scale visual recognition challenge.
\newblock {\em International Journal of Computer Vision}, 115(3), 2015.

\bibitem{shelhamer2016clockwork}
E.~Shelhamer, K.~Rakelly, J.~Hoffman, and T.~Darrell.
\newblock Clockwork convnets for video semantic segmentation.
\newblock In {\em ECCV}, 2016.

\bibitem{shi2011human}
Q.~Shi, L.~Cheng, L.~Wang, and A.~Smola.
\newblock Human action segmentation and recognition using discriminative
  semi-markov models.
\newblock {\em International journal of computer vision}, 93(1):22--32, 2011.

\bibitem{shou2016temporal}
Z.~Shou, D.~Wang, and S.-F. Chang.
\newblock Temporal action localization in untrimmed videos via multi-stage
  cnns.
\newblock In {\em Proceedings of the IEEE Conference on Computer Vision and
  Pattern Recognition}, pages 1049--1058, 2016.

\bibitem{charadeswebsite}
G.~Sigurdsson.
\newblock Charades dataset.
\newblock \url{http://allenai.org/plato/charades/}, 2017.
\newblock Accessed: 2017-04-10.

\bibitem{sigurdsson2016asynchronous}
G.~A. Sigurdsson, S.~Divvala, A.~Farhadi, and A.~Gupta.
\newblock Asynchronous temporal fields for action recognition.
\newblock {\em arXiv preprint arXiv:1612.06371}, 2016.

\bibitem{sigurdsson2016hollywood}
G.~A. Sigurdsson, G.~Varol, X.~Wang, A.~Farhadi, I.~Laptev, and A.~Gupta.
\newblock Hollywood in homes: Crowdsourcing data collection for activity
  understanding.
\newblock In {\em ECCV}, 2016.

\bibitem{simonyan2014two}
K.~Simonyan and A.~Zisserman.
\newblock Two-stream convolutional networks for action recognition in videos.
\newblock In {\em Neural Information Processing Systems}, 2014.

\bibitem{simonyan2014very}
K.~Simonyan and A.~Zisserman.
\newblock Very deep convolutional networks for large-scale image recognition.
\newblock {\em arXiv preprint arXiv:1409.1556}, 2014.

\bibitem{srivastava15_unsup_video}
N.~Srivastava, E.~Mansimov, and R.~Salakhutdinov.
\newblock Unsupervised learning of video representations using {LSTM}s.
\newblock In {\em ICML}, 2015.

\bibitem{thrun2005probabilistic}
S.~Thrun, W.~Burgard, and D.~Fox.
\newblock {\em Probabilistic Robotics (Intelligent Robotics and Autonomous
  Agents)}.
\newblock The MIT Press, 2005.

\bibitem{tran2015learning}
D.~Tran, L.~Bourdev, R.~Fergus, L.~Torresani, and M.~Paluri.
\newblock Learning spatiotemporal features with 3d convolutional networks.
\newblock In {\em International Conference on Computer Vision (ICCV)}, 2015.

\bibitem{varol2016long}
G.~Varol, I.~Laptev, and C.~Schmid.
\newblock Long-term temporal convolutions for action recognition.
\newblock {\em arXiv preprint arXiv:1604.04494}, 2016.

\bibitem{venugopalan2015sequence}
S.~Venugopalan, M.~Rohrbach, J.~Donahue, R.~Mooney, T.~Darrell, and K.~Saenko.
\newblock Sequence to sequence-video to text.
\newblock In {\em International Conference on Computer Vision}, 2015.

\bibitem{vondrick2016anticipating}
C.~Vondrick, H.~Pirsiavash, and A.~Torralba.
\newblock Anticipating visual representations from unlabeled video.
\newblock In {\em Computer Vision and Pattern Recognition (CVPR)}, 2016.

\bibitem{vondrick2016generating}
C.~Vondrick, H.~Pirsiavash, and A.~Torralba.
\newblock Generating videos with scene dynamics.
\newblock In {\em Neural Information Processing Systems (NIPS)}, 2016.

\bibitem{vae_eccv2016}
J.~Walker, C.~Doersch, A.~Gupta, and M.~Hebert.
\newblock An uncertain future: Forecasting from variational autoencoders.
\newblock In {\em European Conference on Computer Vision}, 2016.

\bibitem{wang2013action}
H.~Wang and C.~Schmid.
\newblock Action recognition with improved trajectories.
\newblock In {\em International Conference on Computer Vision}, 2013.

\bibitem{wangaction}
L.~Wang, Y.~Qiao, and X.~Tang.
\newblock Action recognition and detection by combining motion and appearance
  features.
\newblock In {\em THUMOS'14 challenge entry}, 2014.

\bibitem{wang2015towards}
L.~Wang, Y.~Xiong, Z.~Wang, and Y.~Qiao.
\newblock Towards good practices for very deep two-stream convnets.
\newblock {\em arXiv preprint arXiv:1507.02159}, 2015.

\bibitem{wang2016temporal}
L.~Wang, Y.~Xiong, Z.~Wang, Y.~Qiao, D.~Lin, X.~Tang, and L.~Van~Gool.
\newblock Temporal segment networks: towards good practices for deep action
  recognition.
\newblock In {\em European Conference on Computer Vision}, 2016.

\bibitem{wang2006hidden}
S.~B. Wang, A.~Quattoni, L.-P. Morency, D.~Demirdjian, and T.~Darrell.
\newblock Hidden conditional random fields for gesture recognition.
\newblock In {\em Computer Vision and Pattern Recognition (CVPR)}, 2006.

\bibitem{wiskott2002slow}
L.~Wiskott and T.~J. Sejnowski.
\newblock Slow feature analysis: Unsupervised learning of invariances.
\newblock {\em Neural computation}, 14(4):715--770, 2002.

\bibitem{visualdynamics16}
T.~Xue, J.~Wu, K.~L. Bouman, and W.~T. Freeman.
\newblock Visual dynamics: Probabilistic future frame synthesis via cross
  convolutional networks.
\newblock In {\em NIPS}, 2016.

\bibitem{yeung2015every}
S.~Yeung, O.~Russakovsky, N.~Jin, M.~Andriluka, G.~Mori, and L.~Fei-Fei.
\newblock Every moment counts: Dense detailed labeling of actions in complex
  videos.
\newblock {\em arXiv preprint arXiv:1507.05738}, 2015.

\bibitem{yeung2016end}
S.~Yeung, O.~Russakovsky, G.~Mori, and L.~Fei-Fei.
\newblock End-to-end learning of action detection from frame glimpses in
  videos.
\newblock In {\em CVPR}, 2016.

\bibitem{yuanadsc}
J.~Yuan, Y.~Pei, B.~Ni, P.~Moulin, and A.~Kassim.
\newblock Adsc submission at thumos challenge 2015.
\newblock In {\em THUMOS'15 challenge entry}, 2015.

\bibitem{yue2015beyond}
J.~Yue-Hei~Ng, M.~Hausknecht, S.~Vijayanarasimhan, O.~Vinyals, R.~Monga, and
  G.~Toderici.
\newblock Beyond short snippets: Deep networks for video classification.
\newblock In {\em Computer Vision and Pattern Recognition}, 2015.

\bibitem{zeiler2014visualizing}
M.~D. Zeiler and R.~Fergus.
\newblock Visualizing and understanding convolutional networks.
\newblock In {\em European Conference on Computer Vision}, pages 818--833.
  Springer, 2014.

\end{thebibliography}
}

\end{document}